%% file: IEEE_Paper_LaTeX_Template_LETTER_V3.tex
\documentclass[10pt,conference,letterpaper]{IEEEtran}
\usepackage{times,amsmath,epsfig}
\newtheorem{definition}{Definition}
\newtheorem{theorem}{Theorem}
\usepackage{booktabs} 
\usepackage{balance} 
\usepackage{comment}
\usepackage{subcaption}
\usepackage{graphicx}
\usepackage{array}
\usepackage{multirow}
\usepackage[linesnumbered,ruled]{algorithm2e}
\usepackage{algpseudocode}
\usepackage{amssymb}
\usepackage{cuted}
\usepackage{caption}
\usepackage[colorlinks]{hyperref}
\usepackage{url}
\SetCommentSty{mycommfont}

\DeclareMathOperator*{\argmax}{argmax}

\newcommand{\splitatcommas}[1]{%
  \begingroup
  \begingroup\lccode`~=`, \lowercase{\endgroup
    \edef~{\mathchar\the\mathcode`, \penalty0 \noexpand\hspace{0pt plus 1em}}%
  }\mathcode`,="8000 #1%
  \endgroup
}
\usepackage[textsize=tiny, textwidth=1.5cm]{todonotes}
\title{Get Your Workload in Order:\\
Game Theoretic Prioritization of Database Auditing}

\author{%
{Chao Yan{\small $~^{1}$}, Bo Li{\small $~^{2}$}, Yevgeniy Vorobeychik{\small $~^{1,4}$}, Aron Laszka{\small $~^{3}$}, Daniel Fabbri{\small $~^{1,4}$}, Bradley Malin{\small $~^{1,4}$}}%
\vspace{1.6mm}\\
\fontsize{10}{10}\selectfont\itshape
$^{1}$\,Department of Electrical Engineering and Computer Science, Vanderbilt University, TN 37240, USA\\
$^{2}$\,Department of Electrical Engineering and Computer Science, University of California, Berkeley, CA 94720, USA\\
$^{3}$\,Department of Computer Science, University of Houston, TX 77004, USA\\
$^{4}$\,Department of Biomedical Informatics, Vanderbilt University, TN 37240, USA\\
\fontsize{9}{9}\selectfont\ttfamily\upshape
$^{1,4}$\,\{chao.yan, yevgeniy.vorobeychik, daniel.fabbri, b.malin\}@vanderbilt.edu\\
$^{2}$\,crystalboli@berkeley.edu%
~~~ $^{3}$\,alaszka@uh.edu%
\vspace{1.2mm}\\
\fontsize{10}{10}\selectfont\rmfamily\itshape
\fontsize{9}{9}\selectfont\ttfamily\upshape

}

\begin{document}
\maketitle
\begin{abstract}
The quantity of personal data that is collected, stored, and subsequently processed continues to grow at a rapid pace.  Given its potential sensitivity, ensuring privacy protections has become a necessary component of database management.
To enhance protection, a number of mechanisms have been developed, such as audit logging and alert triggers, which notify administrators about suspicious activities that may require investigation.
 However,
 this approach to auditing is limited in several respects. First, the volume of such alerts grows with the size of the database and is often substantially greater than the capabilities of resource-constrained organizations.
 Second, strategic attackers can attempt to disguise their actions or carefully choosing which records they touch, such as by limiting the number of database accesses they commit, thus potentially hiding illicit activity in plain sight.
 In this paper, we introduce a novel approach to database auditing that explicitly accounts for adversarial behavior by
 1) prioritizing the order in which types of alerts are investigated and 2) providing an upper bound on how much resource to allocate for each type.

Specifically, we model the interaction between a database auditor and potential attackers as
a Stackelberg game in which the auditor chooses a (possibly
randomized) auditing policy and attackers choose
which, if any, records to target.
Upon doing so, we show that even a highly constrained version of the auditing problem is NP-Hard.
Based on this finding, 
we introduce an approach that combines linear
programming, column generation, and heuristic search to derive an
auditing policy.
On the synthetic data, we perform an extensive evaluation on both the approximation degree of our solution with the optimal one and the computational magnitude of our approach. The two real datasets, 
1) 1.5 months of audit logs from the electronic medical record system
of Vanderbilt University Medical Center
and 2) a publicly available credit card application
dataset of 1000 records, are used to test the policy-searching performance.
The findings illustrate that our methods produce
high-quality mixed strategies as database audit policies, and our general approach significantly
outperforms non-game-theoretic baselines.  
\end{abstract}
%

\input{samplebody-conf}

\bibliographystyle{IEEEtran}
\bibliography{IEEEexample}
\input{X_appendix.tex}

\end{document}

%% file: samplebody-conf.tex
\input{I_introduction.tex}

\input{III_alertprioritization.tex}

\input{IV_solving.tex}

\input{V_techniqueevaluate.tex}
\input{VI_frameworkevaluate.tex}

\input{II_relatedwork.tex}

\input{VII_discussion.tex}

\input{IX_acknowledge.tex}

%% file: I_introduction.tex
\section{Introduction}
Modern computing and storage technology has made it possible to create \emph{ad hoc} database systems with the ability to collect, store, and process extremely detailed information about the daily activities of individuals \cite{mcafee2012}. These database systems hold great value for society, but accordingly face challenges to security and, ultimately, personal privacy. The sensitive nature of the data stored in such systems attracts malicious attackers who can gain value by disrupting them in various ways (e.g., stealing sensitive information, commandeering computational resources, committing financial fraud, and simply shutting the system down) \cite{Ablon:2014:MCT:2685860}. It is evident that the severity and frequency of attack events continues to grow. Notably, the most recent breach at Equifax led to the exposure of data on 143 million Americans, including credit card numbers, Social Securiy numbers, and other information that could be used for identity theft or other illicit purposes \cite{cnn2017equifax}.  Even more of a concern is that the exploit of the system continued for at least two months before it was discovered.


While complex access control systems have been developed for database management, it has been recognized that in practice no database systems will be impervious to attack \cite{Ten10}. 
As such, prospective technical protections need to be complemented by retrospective auditing mechanisms, a notion that has been well recognized by the database community \cite{kuna2014outlier}. Though audits do not directly prevent attacks in their own right, they may allow for the discovery of breaches that can be followed up on before they escalate to full blown exploits by adversaries originating from beyond, as well as within, an organization.

In the general situation of database management, auditing relies heavily on the performance of a \emph{threat detection and misuse tracking} (TDMT) module, which raises real-time alerts based on the actions committed to a system for further investigation by experts. In general, the alert types are speciafically predefined by the administrator officials in \emph{ad hoc} applications. For instance, in the healthcare domain, organizations are increasingly reliant on electronic medical record (EMR) systems for anytime, anywhere access to a patient's health status \cite{Hsiao2013}. Given the complex and dynamic nature of healthcare, these organizations often grant employees broad access privileges, which increases the potential risk that inside employees illegally exploite the EMR of patients \cite{Gunter11}. To detect when a specific access to a patient's medical record is a potential policy violations, healthcare organizations use various triggers to generate alerts, which can be based on predefined rules (e.g., when an access is made to a designated very important person).  As a consequence, the detected anomalies, which indicate deviations from routine behavior (e.g., when a pediatrician accesses the records of elderly individuals), can be checked by privacy officials \cite{Agrawal:07}.

Although TDMTs are widely deployed in database systems as both detection and deterrence tools, security and privacy have not been sufficiently guaranteed. The utility of TDMT in practice is currently limited by the fact that they often lead to a very large number of alerts, whereas the number of actual violations tends to be quite small. This is particularly problematic because the large quantities of false alarms can easily overwhelm the capacity of the administrative officials who are expected to follow-up on these, but have limited resources at their disposal \cite{Rostad06}. One typical example is the observation from our evaluation dataset: at Vanderbilt University Medical Center, on any workday, the volume of accesses to the EMR system is around 1.8 million, of which more than 30,000 alerts of varying predefined types are generated, which far beyonds the capacity of privacy officials. Therefore, in lieu of an efficient audit functionality in the database systems, TDMTs are not optimized for detecting suspicious behavior. 

Given the overwhelming number of alerts in comparison to available auditing resource and the need to catch attackers, the core query function invoked by an administrator must consider resource constraints. And, given such constraints, we must determine which triggered alerts should be recommended for investigation.  One intuitive way to proceed is to prioritize alert categories based on potential impact of a violation, if one were to be found. However, this is an inadequate strategy because would-be violators can be strategic and, thus, reason about the specific violations they can perform so that they balance the chance of being audited with the benefits of the violation. To address this challenge, we introduce a model based on a Stackelberg game, in which an auditor chooses a randomized auditing policy, while potential violators choose their victims (such as which medical records to view) or to refrain from malicious behavior after observing the auditing policy.

Specifically, our model restricts the space of audit policies to consider two dimensions: 1) how to prioritize alert categories and 2) how much resource to allocate to each category. We show that even a highly restricted version of the auditor's problem is NP-Hard. Nevertheless, we propose a series of algorithmic methods for solving these problems, leveraging a combination of linear programming and column generation to compute an optimal randomized policy for prioritizing alert categories. We perform an extensive experimental evaluation with two real datasets---one involving EMR access alerts and the other pertaining to credit card eligibility decisions---the results of which demonstrate the effectiveness of our approach.

The remainder of the paper is organized as follows. In Section ~\ref{sec:game}, we formally define the game theoretic alert prioritization problem and prove its NP-hardness.
In Section~\ref{sec:solving}, we describe the algorithmic approaches for computing a randomized audit policy.
In Section ~\ref{sec:technique}, we introduce a synthetic dataset to show, in a controlled manner, the effectiveness
of 
our methods for approximating the optimal solution with dramatic gains in efficiency.
In Sections ~\ref{sec:framework}, we use two real datasets (from healthcare and finance) that rely upon predefined alert types
to show that our methods lead to high-quality audit strategies. In Section~\ref{sec:related}, we discuss related work on alert processing in the database systems, security and game theory, and 
audit games.
We discuss our findings and conclude this paper in Section~\ref{sec:discussion}.

%% file: III_alertprioritization.tex
\section{Game Theoretic Model of Alert Prioritization}
\label{sec:game}

In environments dealing with sensitive data or critical services, it is common to deploy
TDMTs to raise alerts upon observing suspicious events.
By defining \emph{ad hoc} alert types, each suspicious event can be marked with an alert label, or type, and put into an audit bin corresponding to this type. 
Typically, the vast majority of the raised alerts do not correspond to actual attacks, as they are generated as a part of routine workflow that is too complex to accurately capture.
Consequently, looking for actual violations amounts to looking for needles in a large haystack of alerts, and inspecting all, or even a large proportion of, alerts that are typically generated is rarely feasible.
A crucial consideration, therefore, is how to \emph{prioritize} alerts, choosing a subset that can be audited given a specified auditing budget from a vast pool of possibilities.
The prioritization problem is complicated by the fact that intelligent adversaries---that is, would-be violators of organizational access policies---would react to an auditing policy by changing their behavior 
to balance the gains from violations, and the likelihood, and consequences, of detection.

We proceed to describe a formal model of alert prioritization as a game between an auditor, who chooses an alert prioritization policy, and multiple attackers, who determine the nature of violation, or are deterred from one, in response. In the described scenarios, we assume that the attackers have complete information.
For reference purposes, the symbols used throughout this paper are described in Table \ref{table_1}.
\begin{table}
	\centering
	\caption{A legend of the notation used in this paper.}
	\label{table_1}
	\begin{tabular}{ll}
		\toprule 
		\hline		
		\bfseries Symbols & \bfseries Interpretation \\
		\midrule
		$T$ & Set of alert types \\
		$\mathbf{E}$ & Set of entities or users causing events\\
		$\mathbf{V}$ & Set of records or files available for access\\
		$P_{ev}^t$ & Probability of raising type $t$ alert by attack $\left \langle e,v \right \rangle$\\
		$C_t$  & Cost for auditing an alert of type $t$ \\
		$B$ & Auditing budget\\
		$F_t(n)$ & Probability that at most $n$ alerts are in type $t$  \\
		$\pmb{O}$ & Set of all alert prioritizations over $T$ \\
		$Z_t$ & Number of alerts under type $t$  \\
		$b_t$ & Budget threshold assigned for auditing type $t$ \\
		$R(\left \langle e,v \right \rangle)$ & Adversary's gain when attack $\left \langle e,v \right \rangle$ is undetected  \\
		$M(\left \langle e,v \right \rangle)$ & Adversary's penalty when attack $\left \langle e,v \right \rangle$ is captured \\
		$K(\left \langle e,v \right \rangle)$ & Cost of deploying attack $\left \langle e,v \right \rangle$ \\
		$p_{\pmb{o}}$ & Probability of choosing an alert prioritization $\pmb{o}$ \\
		$p_e$ & Probability that $e$ is a potential adversary\\
		\bottomrule
	\end{tabular}
\end{table}

\subsection{System Model}

Let $\mathbf{E}$ be the set of potential adversaries, such as employees in a healthcare organization, some of whom could be potential violators of privacy policies, and $\mathbf{V}$ be the set of potential victims, such as patients in a healthcare facility.
We define events, as well as attacks, by a tuple $\langle e,v \rangle$.
A subset of these events will trigger alerts.
Now, let $T$ be the set of alert types, or categorical labels assigned to different kinds of suspicious behavior.
For example, a doctor viewing a record for a patient not assigned to them and a nurse viewing the EMR for another nurse (who is also a patient) in the same healthcare facility could trigger two distinct alert types.
We assume that each event $\langle e,v \rangle$ maps to at most one alert type $t \in T$.
This mapping may be stochastic; that is, given an event $\langle e,v \rangle$, an alert $t$ is triggered with probability $P_{ev}^t$, and no alert is triggered otherwise (i.e., $P_{ev}^{t'} = 0$ for all $t' \ne t$).
Typically, both categorization of alerts and corresponding mapping between events and types is given (for example, through predefined rules).
If not, it can be inferred by generating possible attacks and inspecting how they are categorized by TDMT.
Auditing each alert is time consuming and the time to audit an alert can vary by alert type.
Let $C_t$ be the cost (e.g., time) of auditing a single alert of type $t$ and let $B$ be the total budget allocated for auditing.

Normal events resulting in alerts arrive based on a distribution reflecting a typical workflow of the organization.
We assume this distribution is known, represented by $F_t(n)$, which is the probability that at most $n$ alerts of type $t$ are generated.
If we make the reasonable assumption that attacks are rare events and that the alert logs are tamper-proof by applying certain technique, then this distribution can be obtained from historical alert logs. It is noteworthy that the probability that adversaries successfully manipulate the distribution in the sensitive practices (e.g., the EMR system or the credit card application program), to fool the audit model is almost zero. The cost of orchestrating and implementing such attacks is much higher than what could be gained from running a few undetected attacks.

\subsection{Game Model}

We model the interaction between the auditor and potential violators as a Stackelberg game.
Informally, the auditor chooses a possibly randomized auditing policy, which is observed by the prospective violators, who in response choose the nature of the attack, if any.
Both decisions are made \emph{before the alerts produced through normal workflow are generated} according to a known stochastic process~$F_t(n)$.

In general, a specific pure strategy of the defender (auditor) is a mapping from an arbitrary realization of alert counts of all types to a subset of alerts that are to be inspected, abiding by a constraint on the total amount of budget $B$ allocated for auditing alerts.
Even representing a single such strategy is intractable, let alone optimizing in the space of randomizations over these.
We therefore restrict the defender strategy space in two ways.
First, we let pure strategies involve an ordering $\pmb{o} = (o_1, o_2, \ldots, o_{|T|}) ~~$ ($\forall i,j \in \mathbb{Z}^{+}$ and $i,j \in [1,|T|]$, if $i \ne j$, then $o_i \ne o_j$) over alert types, where the subscript indicates the position in the ordering, and a vector of thresholds $\mathbf{b} = (b_1,\ldots,b_{|T|})$, with $b_t$ being the maximum budget available for auditing alerts in category $t$.
Let $\mathbf{O}$ be the set of feasible orderings, which may be a subset of all possible orders over types (e.g., organizational policy may impose constraints, such as always prioritizing some alert categories over others).
We interpret a threshold $b_t$ as the maximum budget allocated to $t$; thus, the most alerts of type $t$ that can be inspected is $\lfloor b_t / C_t \rfloor$.
Second, we allow the auditor to choose a randomized policy over alert orderings, with $p_{\pmb{o}}$ being the probability that ordering $\pmb{o}$ over alert types is chosen, whereas the thresholds $\mathbf{b}$ are deterministic and independent of the chosen alert priorities.

We have a collection of potential adversaries $\mathbf{E}$, each of whom may target any potential victim $v \in \mathbf{V}$.
We assume that the adversary will target exactly one victim (or at most one, if $\mathbf{V}$ contains an option of not attacking anyone).
Thus, the strategy space of each adversary $e$ is $\mathbf{V}$.
In addition, we assume that any given potential adversary is actually unlikely to consider attacking.
We formalize it by introducing a probability $p_e$ that an attack by $e$ is considered at all (i.e., $e$ does not even consider attacking with probability $1-p_e$).


Suppose we fix a prioritization $\pmb{o}$ and thresholds $\mathbf{b}$.
Let $o(t)$ be the position of alert type $t$ in $\pmb{o}$ and $o_i$ be the alert type in position $i$ in the order.
Let $B_{t}(\pmb{o}, \mathbf{b}, \mathbf{Z})$ be the budget remaining to inspect alerts of type $t$ if the order is $\pmb{o}$, the defender uses alert type thresholds $\mathbf{b}$, and the vector of realizations of benign alert type counts is $\mathbf{Z} = \{Z_1,\ldots,Z_{|T|}\}$.
Then we have
\begin{equation*}
\begin{aligned}
B_{t}(\pmb{o},&  \mathbf{b}, \mathbf{Z}) = \\
&\max \left\{\left\lfloor \left(B - \sum_{i=1}^{o(t)-1}\min\left\{b_{o_i}, Z_{o_i}C_{o_i} \right\}\right)/C_{t}\right\rfloor,0\right\}.
\end{aligned}
\end{equation*}

Now, let us take a moment to unpack this expression for context.
For the audited alert type $t$, we repeatedly compare the threshold $b_{t}$ with $Z_{t} C_{t}$ to determine how much budget will be left for the types that follow in the priority order.
If the total budget that is eaten by inspecting alerts prior to $t$ is larger than $B$, $B_{t}(\pmb{o}, \mathbf{b}, \mathbf{Z})$ returns $0$, and no alerts of type $t$ will be inspected.
Next, we can compute the number of alerts of type $t$ that are audited as
$$n_{t}(\pmb{o},\mathbf{b}, \mathbf{Z}) = \min\left\{B_{t}(\pmb{o},\mathbf{b}, \mathbf{Z}),\left\lfloor b_{t}/C_{t}\right\rfloor ,Z_{t}\right\}.$$
Suppose that an attack generates an alert of type $t$.
As noted earlier, we assume that the number of alerts generated due to attacks is a negligible proportion of all generated alerts (e.g., when $p_e$ are small).
Then, the probability that an alert of type $t$ generated through an attack is detected is approximately

\begin{equation}{\label{prd}}
P_{al}(\pmb{o}, \mathbf{b}, t) \approx \mathbb{E}_{\mathbf{Z}}\left[\frac{n_{t}(\pmb{o},\mathbf{b}, \mathbf{Z)}}{Z_{t}}  \right].
\end{equation}
We can further approximate this probability by sampling from the joint distribution over alert type counts $\mathbf{Z}$.

The adversary $e$ does not directly choose alert types, but rather the victims $v$ (e.g., an EMR).
The probability of detecting an attack $\left \langle e,v \right \rangle$ under audit order $\pmb{o}$ and audit thresholds $\mathbf{b}$ is then
\begin{equation}\label{pr_alert}
P_{at}(\pmb{o}, \mathbf{b},  \left \langle e,v \right \rangle) = \sum_{t} P_{ev}^{t} P_{al}(\pmb{o},\mathbf{b}, t).
\end{equation}


We now have sufficient preliminaries to define the utility functions of the adversaries $e \in \mathbf{E}$.
Let $M(\left \langle e,v \right \rangle)$ denote the penalty of the adversary when captured by the auditor, $R(\left \langle e,v \right \rangle)$ denote the benefit if the adversary is not audited, and $K(\left \langle e,v \right \rangle)$ the cost of an attack.
One natural example is $R(\left \langle e,v \right \rangle) = \sum_t w_t I_t$, where $I_t$ is a Boolean indicator of the presence of alert type $t$ and ${w_t}$ the severity of this alert category.
The utility of the adversary is then
\begin{equation}{\label{att_utility}}
\begin{aligned}
U_a(\pmb{o}, & \mathbf{b},  \left \langle e, v \right \rangle) = P_{at}(\pmb{o}, \mathbf{b},  \left \langle e,v \right \rangle)\cdot M(\left \langle e,v \right \rangle) \\
&+ (1 - P_{at}(\pmb{o}, \mathbf{b},  \left \langle e,v \right \rangle))\cdot R(\left \langle e,v \right \rangle) - K(\left \langle e,v \right \rangle).
\end{aligned}
\end{equation}
We assume that the game is zero-sum.
Thus, the auditor's goal is to find a randomized strategy $p_{\pmb{o}}$ and type-specific thresholds $\mathbf{b}$ 
to minimize the expected utility of the adversary:
\begin{equation}
\label{E:auditProblem}
\min_{p_{\pmb{o}}, \mathbf{b}}\quad \sum_{e \in \mathbf{E}}\sum_{\pmb{o} \in \pmb{O}} p_{\pmb{o}} \max_v U_a(\pmb{o}, \mathbf{b}, \left \langle e,v \right \rangle).
\end{equation}
We call this optimization challenge the \emph{optimal auditing problem (OAP)}.

Since the game is zero-sum, the optimal auditing policy can be computed using the following mathematical program, which directly extends the standard linear programming formulation for computing mixed-strategy Nash equilibria in zero-sum games:
\begin{equation}{\label{basic_game}}
\begin{aligned}
\begin{array}{l}
\min_{ \mathbf{b},p_{\pmb{o}}, \mathbf{u}} ~ \sum_{e \in \mathbf{E}} p_e u_e\\
\;\;\;\;\;\;\;\;\; \\
s.t.\;\;\;\;\; \forall \left \langle e,v \right \rangle , ~~u_e \ge   \sum_{\pmb{o}\in \pmb{O}}p_{\pmb{o}}U_a (\pmb{o}, \mathbf{b}, \left \langle e,v \right \rangle)   \\
\;\;\;\;\;\;\;\;\; \\
\;\;\;\;\;\;\;\;\; \sum_{\pmb{o}\in \pmb{O}}p_{\pmb{o}} = 1,\\
\;\;\;\;\;\;\;\;\; \\
\;\;\;\;\;\;\;\;\; \forall \pmb{o}\in \pmb{O},\;\;\; 0\leq p_{\pmb{o}} \leq 1.\\
\;\;\;\;\;\;\;\;\; \\
\end{array}
\end{aligned}
\end{equation}
Indeed, if we fix the decision variables $\mathbf{b}$, the formulation becomes a linear program.
Nevertheless, since the set of all possible alert prioritizations is exponential, even this linear program has exponentially many variables.
Furthermore, introducing decision variables $\mathbf{b}$ makes it non-linear and non-convex.
Next, we show that solving this problem is NP-hard, even in a restricted special case.
We prove this by reducing from the 0-1 Knapsack problem.

\begin{definition}[0-1 Knapsack Problem]{\label{Knapsack}}
Let $I$ be a set of items where each item $i \in I$ has a weight $w_i$ and a value $v_i$, with $w_i$ and $v_i$ integers.
$W$ is a budget on the total amount of weight (an integer).
Question: given a threshold $K$, does there exist a subset of items $R \subseteq I$ such that $\sum_{i \in R} v_i \ge K$ and $\sum_{i \in R} w_i \le W$?
\end{definition}

\begin{theorem}{\label{nptheorem}}
	OAP is NP-hard even when $\mathbf{O}$ is a singleton. 
\end{theorem}

The proof of this theorem is in the Appendix.

%% file: IV_solving.tex
\section{Solving the alert prioritization game}
\label{sec:solving}

There are two practical challenges that need to be addressed to compute useful approximate solutions to the OAP.
First, there is an exponential set of possible orderings of alert types that needs to be considered to compute an optimal randomized strategy for choosing orderings. Second, there is a combinatorial space of possible choices for the threshold vectors $\mathbf{b}$.
In this section, we develop a column generation approach for the linear program induced when we fix a threshold vector $\mathbf{b}$. 
We then introduce a search algorithm to compute the auditing thresholds.


\subsection{Column Generation Greedy Search}

By fixing $\mathbf{b}$, Equation~\ref{basic_game} becomes a linear program, albeit with an exponential number of variables.
However, since the number of constraints is small, only a limited number of variables will be non-zero. The challenge is in finding this small basis.
We do so using column generation, an approach in which we iteratively solve a linear program with a small subset of variables, and then add new variables with a negative reduced cost. We refer to this method as Column Generation Greedy Search (CGGS), the pseudocode for which is in Algorithm~\ref{alg_cg}.


Specifically, we begin with a small subset of alert prioritizations $\pmb{Q} \subseteq \pmb{O}$. We solve the linear program induced after fixing $\mathbf{b}$ in Equation~\ref{basic_game}, restricted to columns in $\pmb{Q}$. For reference purposes, we call this the \emph{master problem}, which is generated by function $G_{lp}(*)$.
Next, we check if there exists a column (ordering over types) that improves upon the current best solution.
The column of parameter matrix of constraints can be denoted as $\mathbf{\Gamma}_{p_{\pmb{o}}} = P_{at}(\pmb{o}, \mathbf{b},  \left \langle e,v \right \rangle)-\mathbf{1} $ for the decision variable $p_{\pmb{o}}$ or $\mathbf{\Gamma}_{u_e} =\mathbf{1}$ for the decision variable $u_e$. The corresponding reduced costs, computed by function $rc(*)$, are $\mathbf{C^{r}}_{p_{\pmb{o}}} = 1 - \mathbf{\pi}_{\pmb{Q}}\cdot\mathbf{\mathbf{\Gamma}}_{p_{\pmb{o}}} $ and  $\mathbf{C^{r}}_{u_e} = - \mathbf{\pi}_{\pmb{Q}}\cdot\mathbf{\Gamma}_{u_e} $, where $\mathbf{\pi}_{\pmb{Q}}$ is the solution of the dual problem. By minimizing the reduced costs, we generate one new column in each iteration and add it to the subset of columns $\pmb{Q}$ in the master problem. Within the process of generating a new column, we use $\mathbf{\Gamma'_{(\pmb{o'} + t)}}$ to denote the parameter column with the audit order $(\pmb{o'} + t)$. This process is repeated until we can prove that the minimum reduced cost is non-negative. 


The subproblem of generating the best column is itself non-trivial.
We address this subproblem through the application of a greedy algorithm for generating a reduced-cost-minimizing ordering over alert types.
The intuition behind CGGS is that, in the process of generating a new audit order, we greedily add one alert type at a time
to minimize the reduced cost \emph{given the order generated thus far}.
We continue until the objective (reduced cost) fails to improve.


\begin{algorithm}
	\caption{Column Generation Greedy Search (CGGS)}
	\label{alg_cg}
	\SetKwInOut{Input}{Input}
	\SetKwInOut{Output}{Output}
	\Input{The set $\pmb{Q}$ with a single random pure strategy for the auditor.}
	\Output{The set of pure strategies $\pmb{Q}$.}
	\While{\underline{True}}{
		$Z = G_{lp}(\pmb{Q})$;\tcc*[f]{Construct LP using current $\pmb{Q}$}\\
		$\pmb{\pi}_{\pmb{Q}} = LP(Dual(Z))$;\tcc*[f]{Solve dual problem}\\
		$\pmb{o'} = []$\;
		\While{\underline{$|\pmb{o'}|<|T|$}}{
			$\pmb{o'} = \pmb{o'} + \argmax_{t \in T\backslash \pmb{o'} }\pmb{\pi}_{\pmb{Q}} \cdot \mathbf{\Gamma'_{(\pmb{o'} + t)}}$\;
		}
		\eIf{\underline{$\min rc(\pmb{Q}) < 0$}}{
			$\pmb{Q} = \pmb{Q} + \pmb{o'}$\;
		}{
			break\;	
		}
	}

\end{algorithm}

\subsection{Iterative Shrink Heuristic Method}


Armed with an approach for solving the linear program induced by a fixed budget threshold vector $\mathbf{b}$, we now develop a heuristic procedure to find alert type thresholds.

First, it should be recognized that $\sum_t b_t \ge B$ because to allow otherwise would clearly waste auditing resources.
Yet there is no explicit upper bound on the thresholds. However, given the distribution of the number of alerts $Z_t$ for an alert type $t$, we can obtain an approximate upper bound on $b_t$, where $F_t(b_t / C_t) \approx 1$.  This is possible because setting the thresholds above such bounds would lead to negligible improvement.
Consequently, searching for a good solution can begin with a vector of audit thresholds, such that for each $b_t$, $F_t(b_t / C_t) \approx 1$.
Leveraging this intuition, we design an
heuristic method, which iteratively shrinks the values of a good subset of audit thresholds according to
a certain step size $\epsilon$.\footnote{``Good'' in this context means that shrinking thresholds within the subset improves the value of the objective function.} We refer to this as the Iterative Shrink Heuristic Method (ISHM), the pseudocode for which is provided in Algorithm~\ref{alg_heu}.

In each atomic searching action, ISHM first makes a subset of thresholds $b_t$ strategically shrink. Next, it checks to see if this
results in an improved solution. We introduce a variable $l_h$, which indicates the level (or the size) of the given subset of $\mathbf{b}$, and $\epsilon \in (0,1)$, which controls the step size.

At the
beginning, the vector of audit thresholds $\{\hat{H}_o\}$ is initialized with the approximate upper bounds.
Then, by assigning $l_h=1$, we consider
shrinking each of the audit thresholds $\hat{H}_i$. The coefficient for shrinking is defined by the $ratio$ in line $7$, which is instantiated with the predefined step size $\epsilon$; i.e., $i = 1$. If the best value for the objective function in the candidate subsets at $l_h=1$ after shrinking shows an improvement, then the shrink is accepted and
the shrinking coefficient is made smaller by increasing $i$. When no coefficient leads to improvement,
we increase $l_h$ by one, which induces
tests of threshold combinations at the same shrinking ratio. This logic is described in line $6$ through $20$.

Once an improvement occurs, the search course
resets
based on the current $\mathbf{b}$. The
search terminates once $l_h>|T|$.

Note that for a single improvement, the worst-case time complexity is $O(\left \lceil 1/\epsilon \right \rceil\cdot O(LP) \cdot 2^{|T|})$. Though exponential, our experiments show that ISHM achieves outstanding performance, both in terms of precision (of approaching the optimal solution) and efficiency.

\begin{algorithm}
	\caption{Iterative Shrink Heuristic Method (ISHM)}
	\label{alg_heu}
	\SetKwInOut{Input}{Input}
	\SetKwInOut{Output}{Output}
	\Input{Instance of the game, step size $\epsilon$.}
	\Output{Vector of audit thresholds $\{\hat{H_i}\}$.}
	
	Initialize $\{\hat{H}_i\}$ with  full coverages in $\{F_t\}$\;
	$l_h = 1$; $obj = +\infty$\;
	\While{\underline{$l_h <= |T|$}}{
		$C_{l_h} = choose(|T|,l_h)$;\tcc*[f]{Find combinations}\\
		$prgrs = 0$\;
		\For{$i\leftarrow 1$ \KwTo $\left \lceil 1/\epsilon \right \rceil$}{
			$ratio = \max\{0,1-i*\epsilon \}$\;
			$obj_r = +\infty$; $pst_r = 0$\;
			
			\For{$j\leftarrow 1$ \KwTo $ |C_{l_h}|$}{
				$temp =  \{\hat{H}_i\}$\;
				\For{$k\leftarrow 1$ \KwTo $l_h$}{
					$temp(1,C_{l_h(j,k)}) ~~ *= ~~ratio$\;
				}
				$ obj' = LP(B,temp)$\; \tcc*[f]{Return LP objective value}\\
				\lIf{\underline{$obj' < obj_r$}}{$obj_r = obj'$; $pst_r = j$}
			}
			\If{\underline{$obj_r < obj$}}{
				$obj = obj_r$\;
				$S_u = C_{l_h}(pst_r,:)$\;\tcc*[f]{Types in need of update}\\
				\lFor{$j\leftarrow 1$ \KwTo $ |S_u|$}{$\hat{H}_{S_{u_j}}~~ *= ~~ratio$}
				break\;
			}
			$prgrs = i$\;
		}
		\lIf{\underline{$prgrs == \left \lceil 1/\epsilon \right \rceil $}}{$l_h ~~+= ~~1$} \lElse{$l_h = 1$}
	}
\end{algorithm}

%% file: V_techniqueevaluate.tex
\section{CONTROLLED EVALUATION}
\label{sec:technique}
To gain intuition into the potential for our methods, we evaluated the performance of the
ISHM and
CGGS approaches using a synthetic dataset,
\emph{Syn\_A}. To enable comparison with an optimal solution, we use a relatively small synthetic dataset, but as will be clear, it is sufficient to illustrate the relationship between our methods and the optimal brute force solution.

To perform the analysis,
we vary the audit budgets $B$ and step size
$\epsilon$ of ISHM.
In addition, we evaluate a combination of CGGS+ISHM (since the
former is also an approximation), by again comparing to the optimal.

\subsection{Data Overview}

The dataset \emph{Syn\_A} consists of $5$ potential attackers who
perform accesses ($p_e = 1$\footnote{The artificially high incidence
  of attacks here is merely to facilitate a comparison with a
  brute-force approach.}), $8$ files, $4$ predefined alert types, and
a set of rules for triggering alerts if any access happens. Table
\ref{wrap-tab1} summarizes the information of \emph{Syn\_A} and
related parameters in the corresponding scenario. We let the number of
alerts for all types be distributed according to a Gaussian
distribution with means and standard deviation as reported in Table~\ref{wrap-tab:1a}.
Since the number of alerts
are integers, we discretize
the $x$-axis of each
alerts cumulative distribution function and use the corresponding
probabilities for each possible alert count.
We consider the $99.5\%$ probability coverage for each alert type to
obtain a finite upper bound on alert counts.

We assume alerts are triggered
deterministically for each access, a common case in rule-based systems.
 The alert type that will be triggered for each potential access is provided in Table \ref{wrap-tab:1b}, where ``-'' represents a benign access. This table is generated with a probability vector $[0.07,0.38,0.23,0.16,0.16]$ for each employee, which corresponds to alert type vector $[0,1,2,3,4]$. Although in reality, benign accesses may be more frequent, we lower their probability to better differentiate the final value of the objective function. The benefit of the adversary for a successful attack, the cost of an attack and the cost of an audit are all directly related to the alert type, which are shown in Table \ref{wrap-tab:1a}. In addition, the penalty for being caught is set to a constant value of $4$.

\begin{table}[h]
	\caption{Description of Dataset \emph{Syn\_A}.}
	\begin{subtable}{0.5\textwidth}
		\caption{Parameter values for alert types in the synthetic setting.}\label{wrap-tab:1a}
		\centering
		\begin{tabular}{ccccc}
			\toprule
			\hline
			\textbf{ }  & \textbf{Type 1}  & \textbf{Type 2} & \textbf{Type 3}  & \textbf{Type 4} \\\hline
			\textbf{ Mean} & 6 & 5  & 4 & 4 \\
			\textbf{ Std} & 2 & 1.6 & 1.3 & 1 \\
			\textbf{ 99.5\% Coverage} & +/-5 & +/-4  & +/-3 & +/-3 \\
			\textbf{ Benefit} & 3.4 & 3.7  & 4 & 4.3 \\
			\textbf{ Attack Cost} & 0.4 & 0.4  & 0.4 & 0.4 \\
			\textbf{ Audit Cost} & 1 & 1  & 1 & 1 \\ \hline
		\end{tabular}
		
	\end{subtable}
	\newline
	\vspace*{0.5 cm}
	\newline
	\begin{subtable}{0.5\textwidth}
		\caption{Rules for alert types in the synthetic setting.}\label{wrap-tab:1b}
		\centering
		\begin{tabular}{@{} c*{8}{>{$}c<{$}} @{}}
			\toprule
			\hline			
			\multirow{2}{*}{\textbf{Employee}} & \multicolumn{7}{c@{}}{\textbf{Record}}\\[-0.1em]
			\cmidrule(l){2-9}
			 &  \textbf{$r_1$}  & \textbf{$r_2$}  & \textbf{$r_3$} & \textbf{$r_4$} & \textbf{$r_5$} & \textbf{$r_6$} & \textbf{$r_7$} & \textbf{$r_8$} \\[-0.1em]
			\midrule
			\textbf{$e_1$}  & - & 3 & 2 & 2 & 3 & 4 & 3 & 1 \\[-0.1em]
	
			\textbf{$e_2$}  & 1 & - & 1 & 1 & 1 & 2 & 1 & 1 \\[-0.1em]
	
			\textbf{$e_3$}  & 1 & 3 & 4 & - & 1 & 3 & 1 & 4 \\[-0.1em]
		
			\textbf{$e_4$}  & 2 & 1 & 3 & 1 & 4 & 4 & 2 & 2 \\[-0.1em]
		
			\textbf{$e_5$}  & 2 & 3 & 1 & 4 & 2 & 1 & 3 & 2 \\[-0.1em]
			\bottomrule
		\end{tabular}
	
	\end{subtable}

	\label{wrap-tab1}
\end{table}

\subsection{Optimal Solution with Varying Budget}

Based on the given information, we can compute the optimal OAP solution. First, the search space for audit thresholds in this scenario is as follows: 1) for each alert type, the audit threshold $b_t \in \mathbf{N}$, 2) the sum of thresholds for all alert types should be greater than or equal to $B$, 3) for each type, the upper bound of the audit threshold $b_t$ is where $F_t(b_t / C_t) \approx 1$. Concretely, we set vector $\mathit{J} = Mean+|99.5\% Coverage|$ as the upper bound for finding the optimal solution. Thus, the space of the investigation of the optimal solution is $O(\prod_{i=1}^{|T|}(\mathit{J_i}+1))$. Note that $0$ is also a possible choice, which means the auditor will not check the corresponding alert type. Thus, it is infeasible to directly solve the OAP in the instances with a large number of alert types or large $\mathit{J_i}$.

To investigate the performance of the proposed audit model, we
allocated a vector of audit budgets  $\mathbf{B} = \{2, 4, 6, 8, 10,
12, 14, 16, 18, 20\}$, which has a wide range with respect to the
scale of the means of the alert types.
We then apply a brute force search to discover an optimal vector of budget
thresholds for each type.
Table \ref{wrap-tab:4} shows the
optimal solution of OAP for each candidate $\mathbf{B}$, including the
optimal value of the objective function, optimal threshold (using the
smallest optimal threshold whenever the optimal solution is not unique), pure
strategies in the support of the optimal mixed strategy,
and the optimal mixed strategy of the auditor. As
expected, it can be seen that as the
budget increases, the optimal value of the objective function (minimized by the auditor) decreases monotonically.
\begin{table*}
	
	\centering
	\caption{The optimal solution for the auditor under various budgets.}\label{wrap-tab:4}
	\begin{tabular}{cccccc}
		\toprule
		\hline
		
		\textbf{ID} &\textbf{Budget} & \textbf{Optimal Objective Value} & \textbf{Optimal Threshold} & \textbf{Effective Pure Strategy} & \textbf{Optimal Mixed Strategy} \\[-0.1em]
		\midrule
		1 & 2 &  12.2945 & [1,1,1,1] & [2,3,4,1][4,1,3,2][4,2,3,1][4,3,2,1] & [0.3566, 0.3780, 0.1210, 0.1444]  \\[-0.1em]
		
		2 & 4 & 7.7176 & [2,1,1,2] & [1,2,3,4][2,1,3,4][4,2,1,3][4,2,3,1] & [0.4664, 0.0052, 0.0934, 0.4350]\\[-0.1em]
		
		3 & 6 & 3.2651 & [2,2,2,2] & [2,1,3,4][4,1,3,2][4,2,1,3][4,2,3,1] & [0.2748, 0.2341, 0.3293, 0.1618] \\[-0.1em]
		
		4 & 8 & -0.4517 & [3,3,2,2] & [2,1,3,4][4,1,3,2][4,2,1,3][4,2,3,1] & [0.0762, 0.4600, 0.1329, 0.3309]\\[-0.1em]
		
		5 & 10 & -2.1314 & [3,3,3,3] & [1,2,3,4][1,4,3,2][4,1,2,3][4,1,3,2] & [0.3926, 0.0788, 0.4080, 0.1206] \\[-0.1em]
		
		6 & 12 & -3.7345 & [4,4,3,3] & [2,1,3,4][4,2,3,1][4,2,1,3][4,1,3,2] & [0.2028, 0.1554, 0.2076, 0.4342] \\[-0.1em]
		
		7 & 14 & -5.1645 & [5,4,3,3] & [2,1,3,4][4,2,3,1][4,2,1,3][4,1,3,2] & [0.3559, 0.2199, 0.3176, 0.1066] \\[-0.1em]
		
		8 & 16 & -6.4510 & [6,5,4,4] & [2,1,3,4][4,1,3,2][4,2,1,3][4,2,3,1]  & [0.2431, 0.2636, 0.1728, 0.3205]\\[-0.1em]
		
		9 & 18 & -7.4649 & [7,6,5,5] & [2,1,3,4][4,1,3,2][4,2,1,3][4,2,3,1] & [0.2710, 0.2630, 0.2054, 0.2615] \\[-0.1em]
		
		10 & 20 & -8.1561 & [9,7,6,6] & [1,2,3,4][4,1,2,3][4,1,3,2][4,2,3,1] & [0.2398, 0.1742, 0.2275, 0.3585] \\[-0.1em]
		
		\bottomrule
	\end{tabular}
\end{table*}

\subsection{Findings}
\label{sec:ishmandcggs}

Our heuristic methods aim to find an approximate solution through major reductions in computation complexity.
 In this respect, the search step size $\epsilon$ is a key factor to consider because it could lead the search into a locally optimal solution. To investigate the gap between the objective function with the optimal solution, as well as the influence of $\epsilon$ on the gap, we performed experiments with a series of step sizes $\mathbf{\epsilon}=[0.05, 0.1, 0.15, 0.2, 0.25, 0.3, 0.35, 0.4, 0.45, 0.5]$. Tables \ref{wrap-tab:5} and \ref{wrap-tab:6},
 summarize the results,
  where each cell consists of two items: 1) the minimized sum of the
  maximal utilities of all adversaries obtained using the heuristic method and 2) the corresponding audit threshold vector.

There are three findings worth highlighting. First, when $\epsilon$ is fixed, the approximated values of the objective function decrease as the budget increases. This is akin to the trend shown in Table \ref{wrap-tab:4}. Second, when the budget $B$ is fixed, the approximated values
achieved through ISHM and ISHM+CGGS exhibit a general growth trend as
$\epsilon$ increases. This occurs because larger shrink ratios
increase the likelihood that the heuristic search will miss more of
the good approximate solutions. Third, we find that the ISHM and ISHM+CGGS solutions are close to the optimal.
To
measure the
solution quality as a function of $\epsilon$, we use
$\gamma_{\epsilon} =\frac{1}{|\mathbf{B}|} \sum_i^{|\mathbf{B}|}
|\hat{S}_{\mathbf{B}_i,\epsilon}-S_{\mathbf{B}_i,\epsilon}|/|S_{\mathbf{B}_i,\epsilon}|$,
where $\hat{S}_{\mathbf{B}_i,\epsilon}$ denotes the approximate
optimal values in Tables \ref{wrap-tab:5} and \ref{wrap-tab:6} and
$S_{\mathbf{B}_i,\epsilon}$ denotes the optimal values provided by
Table \ref{wrap-tab:4}.

In Table \ref{wrap-tab:8}, it can be seen that ISHM (and solving the linear program to optimality) achieves solutions near $99\%$ of the optimal (as denoted
by $\gamma^1_{\epsilon}$) when the step size $\epsilon \le 0.2$. Even
the approximately optimal solutions with $\epsilon=0.5$ has a good
approximation ratio (above $89\%$).
As such, it appears that if we choose an appropriate $\epsilon$, then ISHM can perform well.

When we combine ISHM+CGGS (denoted by $\gamma^2_{\epsilon}$),
the approximation quality drops compared to
$\gamma^1_{\epsilon}$, as we would expect, with the lone exception of ($\epsilon =
0.4$).
However, $\gamma^2_{\epsilon}$ is very close to
$\gamma^1_{\epsilon}$, which suggests that our approximate column
generation method does not significantly degrade the quality of the solution.

Next, we consider the computational burden for ISHM to achieve
an approximate target of the optimal solution.
Table \ref{wrap-tab:9} provides the values of the threshold vectors
under various $B$ and $\epsilon$. It can be seen that the number of
threshold candidates explored decreases as the step size grows.
 For a given $\epsilon$, the number of thresholds considered by the
 algorithm initially increases, but then drops as the audit budget
 increases.
The reason that less effort is necessary at the extremes of the budget range is that
 the restart of the test for a single alert type (to find a better
 position) is invoked less frequently.
By contrast, a larger amount of effort is required in the middle of
the budget range due to more frequent restarts (although this yields only a small improvement).

Finally, we investigate the average number for the threshold vectors
explored by the algorithm over the budget range $\mathbf{B}$. For
the various step sizes, we represent the results in vector form
$\mathcal{T} = [403,223,156,121,93,86,68,66,61,47]$. Dividing by the
number of investigations needed to discover the optimal solution, the
resulting ratio vector is $\splitatcommas{ \mathcal{T}' = [0.0831,
  0.0460 ,0.0321, 0.0251, 0.0198, 0.0190, 0.0163, 0.0182, 0.0206,
  0.0210]}$.
Thus, when $\epsilon = 0.2$ (when both
$\gamma^1_{\epsilon}$ and $\gamma^2_{\epsilon}$ are greater than
$0.99$), the number of thresholds explored is only $2.51\%$ of the
entire space. As such, by applying ISHM, the number of investigated
threshold candidates can be greatly reduced without significantly
sacrificing solution quality.

\begin{table*}[htbp]
	
	\centering
	\caption{The approximation of the optimal solutions obtained by ISHM at various levels of $B$ and $\epsilon$.}\label{wrap-tab:5}
	\begin{tabular}{@{} l*{12}{>{$}l<{$}} @{}}
		\toprule
		\hline
		\multirow{2}{*}{$B$} & \multicolumn{10}{c@{}}{\textbf{Approximation of Optimal Loss of the Auditor and corresponding thresholds by ISHM}}\\[-0.1em]
		\cmidrule(l){2-11}
		~ & \epsilon = 0.05  & \epsilon = 0.10  & \epsilon = 0.15 & \epsilon = 0.20 & \epsilon = 0.25 & \epsilon = 0.30 & \epsilon = 0.35 & \epsilon = 0.40 & \epsilon = 0.45 & \epsilon = 0.50 \\[-0.1em]
		\hline
		\multirow{2}{*}{2}  & 12.2945 & 12.2945 & 12.2958 & 12.2945  & 12.2958 & 12.3675 & 12.3675 & 12.2945 & 12.3675 & 12.3675 \\  & [10,1,1,1] & [9,1,1,1] & [9,9,1,1] & [8,1,1,1]  & [8,9,1,1]  &  [7,9,7,7] & [7,9,7,7] & [6,1,1,1] & [6,9,7,7] & [5,9,7,7]  \\[-0.1em]
		
		\multirow{2}{*}{4}  & 7.7176 &  7.7176 & 7.7176 & 7.7176  & 7.7176  & 7.7176 & 7.7181 & 7.8402 & 7.8402 & 7.9037  \\  & [2,1,1,2] & [2,1,1,2] & [2,1,1,2]  & [2,1,1,2]  & [2,1,1,2]  & [2,1,1,2] & [2,1,7,2] & [1,1,7,7] & [1,9,1,3] & [11,9,1,3] \\[-0.1em]
		
		\multirow{2}{*}{6}  & 3.2651 &  3.2651 &  3.2651  &  3.2651 & 3.2651 & 3.2651 & 3.3267 & 3.2744 & 3.4549 & 3.4549 \\  & [2,2,2,2] & [2,2,2,2] &  [2,2,2,2] & [2,2,2,2] &  [2,2,2,2] & [2,2,2,2] & [3,3,2,2] & [2,3,2,2] & [11,2,3,3] & [11,2,3,3] \\[-0.1em]
		
		\multirow{2}{*}{8}  & -0.4517 & -0.4517 & -0.4517  & -0.4517  &-0.4517 & -0.3508 & -0.4517 & -0.4116 & -0.3730 & -0.2910  \\  & [3,3,2,2] & [3,3,2,2] &  [3,3,2,2] & [3,3,2,2]  &  [3,3,2,2] & [4,4,2,2] & [3,3,2,2] & [11,3,2,2] & [3,4,3,3] & [5,4,3,3] \\[-0.1em]
		
		\multirow{2}{*}{10}  & -2.1314 & -2.1314 & -2.1314  & -2.1314  & -2.1314 & -1.9693 & -1.9996 & -2.0119 & -2.0755 & -2.0037  \\  & [3,3,3,3] & [3,3,3,3] & [3,3,3,3]  & [3,3,3,3]  & [3,3,3,3] & [4,4,4,4] & [4,3,4,4] & [3,3,4,4] & [3,4,3,3] & [5,4,3,3] \\[-0.1em]
		
		\multirow{2}{*}{12}  & -3.7345 & -3.7345 & -3.7345  & -3.7345  & -3.7345 & -3.5991 & -3.5627 & -3.4854 & -3.6533 & -3.6873  \\  & [4,4,3,3] & [4,4,3,3] & [4,4,3,3] & [4,4,3,3]  & [4,4,3,3]  & [4,4,4,4] & [4,5,4,4] & [6,5,4,4] & [6,4,3,3] & [5,4,3,3] \\[-0.1em]
		
		\multirow{2}{*}{14}  & -5.0713 & -5.0713 & -5.0430  & -5.0430 & -5.0713 & -5.0962 & -5.0350 & -5.0629 & -5.0713 & -5.0713 \\  & [9,4,3,5] & [9,4,3,5] & [11,5,3,3] & [11,5,3,3] & [5,4,3,5]  & [7,4,4,4]  & [7,5,4,4] & [6,5,4,4] & [6,4,3,7] & [5,4,3,7] \\[-0.1em]
		
		\multirow{2}{*}{16}  & -6.4510 & -6.4510 & -6.4363  & -6.4510 & -6.3823 & -6.4135 & -6.4363 & -6.4510  & -6.3225 & -6.1149  \\  & [6,5,4,4] & [6,5,4,4] & [7,5,4,4] & [6,5,4,4]  & [6,6,5,5]  & [7,6,4,4] & [7,5,4,4] & [6,5,4,4] & [6,9,7,7] & [5,9,7,7] \\[-0.1em]
		
		\multirow{2}{*}{18}  & -7.4649 & -7.4649 & -7.4600  & -7.4490  &-7.4585 & -7.4490 & -7.4320 &  -7.3956 & -7.3612 & -6.1149  \\  & [7,6,5,5] & [7,6,5,5] & [7,7,5,5]  & [8,7,5,5] & [8,6,5,5]  & [7,6,7,7] & [7,9,7,7] & [11,9,7,4] & [6,9,7,7] & [5,9,7,7] \\[-0.1em]
		
		\multirow{2}{*}{20}  & -8.1561 & -8.1561 & -8.1548  & -8.1523  &-8.1520 & -8.1308 & -8.1138 &   -7.6619 &  -7.3612 & -6.1149  \\  & [9,7,6,6] & [9,7,6,6] & [9,7,7,7] & [8,7,7,7]  & [8,9,7,7]  & [11,6,7,7] & [7,9,7,7] & [11,9,7,4] & [6,9,7,7] & [5,9,7,7] \\[-0.1em]

		\bottomrule
	\end{tabular}
	
\end{table*}
\begin{table*}[htbp]
	
	\centering
	\caption{The approximation of the optimal solutions obtained by ISHM + CGGS at various levels of $B$ and $\epsilon$.}\label{wrap-tab:6}
	\begin{tabular}{@{} l*{12}{>{$}l<{$}} @{}}
		\toprule
		\hline
		\multirow{2}{*}{$B$} & \multicolumn{10}{c@{}}{\textbf{Approximation of Optimal Loss of the Auditor and corresponding thresholds by ISHM + CGGS}}\\[-0.1em]
		\cmidrule(l){2-11}
		~ & \epsilon = 0.05  & \epsilon = 0.10  & \epsilon = 0.15 & \epsilon = 0.20 & \epsilon = 0.25 & \epsilon = 0.30 & \epsilon = 0.35 & \epsilon = 0.40 & \epsilon = 0.45 & \epsilon = 0.50 \\[-0.1em]
		\hline
		\multirow{2}{*}{2}  & 12.2967 & 12.2967 & 12.3096 & 12.2967  & 12.3096 & 12.3677 & 12.3677 & 12.2967 & 12.3677 & 12.3677 \\  & [1,1,1,1] & [1,1,1,1] & [9,9,1,1] & [1,1,1,1]  & [8,9,1,1]  &  [7,9,7,7] & [7,9,7,7] & [1,1,1,1] & [6,9,7,7] & [5,9,7,7]  \\[-0.1em]
		
		\multirow{2}{*}{4}  & 7.7214 &  7.7214 & 7.7346 & 7.7214  & 7.7346  & 7.7346 & 7.7346 & 7.9151 & 7.8402 & 7.9045  \\  & [2,1,1,2] & [2,1,1,2] & [2,9,1,2]  & [2,1,1,2]  & [2,9,1,2]  & [2,9,1,2] & [2,9,1,2] & [1,1,1,7] & [1,9,1,3] & [11,9,1,3] \\[-0.1em]
		
		\multirow{2}{*}{6}  & 3.2755 &  3.2755 &  3.2755  &  3.2755 & 3.2755 & 3.2755 & 3.3628 & 3.3267 & 3.4897 & 3.3099 \\  & [2,2,2,2] & [2,2,2,2] &  [2,2,2,2] & [2,2,2,2] &  [2,2,2,2] & [2,2,2,2] & [3,3,2,2] & [2,3,2,2] & [11,2,3,3] & [2,2,3,3] \\[-0.1em]
		
		\multirow{2}{*}{8}  & -0.4422 & -0.4422 & -0.4422  & -0.4422  &-0.2761 & -0.3300 & -0.4006 & -0.4422 & -0.3404 & -0.2761  \\  & [3,3,2,2] & [3,3,2,2] &  [3,3,2,2] & [3,3,2,2]  &  [5,2,2,7] & [4,4,2,2] & [4,3,2,2] & [3,3,2,2] & [3,4,3,3] & [5,2,3,3] \\[-0.1em]
		
		\multirow{2}{*}{10}  & -2.1203 & -2.1203 & -2.1203  & -2.1203  & -2.1203 & -1.9503 & -1.9873 & -2.0091 & -2.0612 & -1.9508  \\  & [3,3,3,3] & [3,3,3,3] & [3,3,3,3]  & [3,3,3,3]  & [3,3,3,3] & [4,4,4,4] & [4,3,4,4] & [3,3,4,4] & [3,4,3,3] & [5,4,3,3] \\[-0.1em]
		
		\multirow{2}{*}{12}  & -3.7215 & -3.7215 & -3.7215  & -3.7215  & -3.7215 & -3.5832 & -3.5448 & -3.4326 & -3.6383 & -3.6768  \\  & [4,4,3,3] & [4,4,3,3] & [4,4,3,3] & [4,4,3,3]  & [4,4,3,3]  & [4,4,4,4] & [4,5,4,4] & [6,5,4,4] & [6,4,3,3] & [5,4,3,3] \\[-0.1em]
		
		\multirow{2}{*}{14}  & -5.0709 & -5.1529 & -5.0430  & -5.0700 & -5.0698 & -5.0857 & -5.0125 & -5.0494 & -5.0698 & -5.0706  \\  & [5,9,3,4] & [5,4,4,4] & [9,4,3,3] & [6,5,3,4]  & [6,4,3,7]  & [7,4,4,4] & [7,5,4,4] & [6,5,4,4] & [6,4,3,7] & [5,4,3,7] \\[-0.1em]
		
		\multirow{2}{*}{16}  & -6.4394 & -6.4394 & -6.4258  & -6.4394 & -6.3683 & -6.4008 & -6.4258 & -6.4394  & -6.3038 & -6.1149  \\  & [6,5,4,4] & [6,5,4,4] & [7,5,4,4]  & [6,5,4,4]  & [6,6,5,5]  & [7,6,4,4] & [7,5,4,4] & [6,5,4,4] & [6,9,7,7] & [5,9,7,7] \\[-0.1em]
		
		\multirow{2}{*}{18}  & -7.4524 & -7.4524 & -7.4465  & -7.4363  &-7.4472 & -7.4359 & -7.4171 &  -7.3825 & -7.3612 & -6.1149  \\  & [7,6,5,5] & [7,6,5,5] & [7,7,5,5]  & [8,7,5,5] & [8,6,5,5]  & [7,6,7,7] & [7,9,7,7] & [11,5,7,7] & [6,9,7,7] & [5,9,7,7] \\[-0.1em]
		
		\multirow{2}{*}{20}  & -8.1448 & -8.1448 & -8.1433  & -8.1398  &-8.1388 & -8.1207 & -8.1043 &   -7.6619 &  -7.3612 & -6.1149  \\  & [9,7,6,6] & [9,7,6,6] & [9,7,7,7] & [8,7,7,7]  & [8,9,7,7]  & [11,6,7,7] & [7,9,7,7] & [11,9,7,4] & [6,9,7,7] & [5,9,7,7] \\[-0.1em]

		\bottomrule
	\end{tabular}
	
\end{table*}

{\setlength{\tabcolsep}{0.1em}
	\begin{table}[h]
		\fontsize{7}{7}\selectfont
		\centering
		\caption{The average precision over the budget vector $\mathbf{B}$ by applying ISHM and ISHM+CGGS.}\label{wrap-tab:8}
		\begin{tabular}{ccccccccccc}
			\toprule
			\hline
			\\[-0.6em]
			$\epsilon$ & $0.05$& $0.10$& $0.15$& $0.20$& $0.25$& $0.30$& $0.35$& $0.40$& $0.45$& $0.50$ \\[0em]
			\midrule
			$\gamma^1_{\epsilon}$ & $0.9982$& $0.9982$& $0.9973$& $0.9974$ & $0.9970$ & $0.9634$ & $0.9830$ & $0.9680$ & $0.9549$ & $0.8982$ \\[0.1em]
			
			$\gamma^2_{\epsilon}$ & $0.9943$& $0.9959$& $0.9932$& $0.9940$ & $0.9560$ & $0.9562$ & $0.9684$ & $0.9700$ & $0.9452$ & $0.8966$ \\[0em]
			
			\bottomrule
		\end{tabular}
		
	\end{table}}

\begin{table}[h]
	\fontsize{7}{7}\selectfont
	\centering
	
	\caption{The number of threshold vectors checked by ISHM with a given budget $B$ and step size $\epsilon$.}\label{wrap-tab:9}
	\begin{tabular}{@{} c*{11}{>{$}c<{$}} @{}}
		\toprule
		\hline
		\\[-0.6em]
		\multirow{2}{*}{$\epsilon$} & \multicolumn{10}{c@{}}{B}\\[0.1em]
		\cmidrule(l){2-11}
		~ & $2$ & $4$ & $6$ & $8$ & $10$ & $12$ & $14$ & $16$ & $18$ & $20$ \\[0.1em]
		\hline
		\\[-0.6em]
		$0.10$ & $251$ & $267$ & $255$ & $243$ & $235$ & $227$ & $199$ & $207$ & $191$ & $171$ \\[0.1em]
		$0.20$ & $128$ & $144$ & $148$ & $140$ & $132$ & $124$ & $108$ & $108$ & $92$ & $84$ \\[0.1em]
		$0.30$ & $65$ & $109$ & $101$ & $93$ & $85$ & $85$ & $81$ & $77$ & $69$ & $65$ \\[0.1em]
		$0.40$ & $74$ & $66$ & $78$ & $70$ & $70$ & $62$ & $62$ & $62$ & $50$ & $50$ \\[0.1em]
		$0.50$ & $35$ & $43$ & $47$ & $47$ & $47$ & $47$ & $43$ & $35$ & $35$ & $35$ \\[0.1em]
		\bottomrule
	\end{tabular}
	
\end{table}

%% file: VI_frameworkevaluate.tex
\section{Model Evaluation}
\label{sec:framework}
The previous results suggest ISHM and CGGS can be efficient and effective in solving the OAP in a small controlled environment. 
Here, we investigate the performance of the proposed game-theoretical audit model on more realistic and larger datasets.
This evaluation consists of comparing the quality of solutions of OAP with several natural alternative auditing strategies.

The first 
dataset, \emph{Rea\_A}, corresponds to the EMR access logs of Vanderbilt University Medical Center (VUMC). This dataset is notable because VUMC privacy officers rely on this data to conduct retrospective audits to determine if there are accesses that violate organizational policy.
The central goal in this use case is to preserve patient privacy.
The second dataset, \emph{Rea\_B}, consists of public observations of credit card applications. It labels applicants as having either low or high risk of fraud.
We provide an audit mechanism to capture events of credit card fraud based on the features in this dataset.

\subsection{Data Overview}
{\bf{\emph{Rea\_A}}} consists of the VUMC EMR access logs for 28 continuous workdays during 2017. There are $48.6M$ access events, $38.7M$ ($79.5\%$) of which are repeated accesses.\footnote{We define a repeated access as an access that is committed by the same employee to the same patient's EMR on the same day.} We filtered out the repeated accesses to focus on the distinct user-patient relationships established on a daily basis. The mean and standard deviation of daily
access events was $355,\!602.18$ and $195,\!144.99$, respectively.
The features for each event include: 1) timestamp, 2) patient ID, 3) employee ID, 4) patient's residential address, 5) employee's residential address, 6) employee's VUMC department affiliation, and 6) indication of if patient is an employee.
We focus on the following alert types:
1) employee and patient share the same last name, 2) employee and patient work in the same VUMC department, 3) employee and patient share the same residential address, and 4) employee and patient are neighbors within a distance threshold.

In certain cases, the same access may generate multiple alerts, each with a distinct type.
For example, if a husband, who is a BMRC employee, accesses his wife's EMR, then two alert types may be triggered: 1 (same last name) and 3 (same address).
We therefore
redefine the set of alert types to also consider combinations of alert categories.
The resulting set of alert types is detailed in Table \ref{wrap-tab:10}.
\begin{table}
	\fontsize{7.5}{7.5}\selectfont
	\centering
	\caption{Description of the EMR alert types.}\label{wrap-tab:10}
	\begin{tabular}{llcccccc}
		\toprule
		\hline
		\\[-0.6em]
		\textbf{ID}  & \textbf{Alert Type Description}  & \textbf{Mean}  & \textbf{Std}  \\\hline
		\\[-0.6em]
		1 & Same Last Name & 183.21 & 46.40   \\[0.1em]
		2 & Department Co-worker &  32.18 & 23.14   \\[0.1em]
		3 & Neighbor ($\le$ 0.5 miles) &  113.89 & 80.44   \\[0.1em]
		4 & Last Name; Same address & 15.43  & 14.61   \\[0.1em]
		5 & Last Name; Neighbor ($\le$ 0.5 miles) & 23.75  & 11.07   \\[0.1em]
		6 & Same address; Neighbor ($\le$ 0.5 miles) & 20.07 & 11.49   \\[0.1em]
		7 & Last Name; Same address; Neighbor ($\le$ 0.5 miles) & 32.07  & 16.54   \\[0.1em]
		\hline
	\end{tabular}
\end{table}

We label each access event in the logs with a corresponding alert type 
or as
``benign'' (i.e., no alerts generated).
To evaluate our methods, we choose a random sample of $50$ employees and patients who generate at least
one alert.
This set of employees and the set of patients then results in $2500$ \emph{potential} accesses, where each employee can access each patient.

We let the probability that an employee could be malicious be $1$, which is artificially high, but enables us to clearly compare the methods. The benefit vector for the adversary is $[10,12,12,24,25,25,27]$ for the corresponding categories of alert types (1-7 in Table~\ref{wrap-tab:10}).
The penalty for capture is set to $15$. We set the cost of both an attack and an audit to $1$.
We acknowledge that the model parameters are \emph{ad hoc}, but this does not affect the results of our comparative analysis. In practice, this would be accomplished based on expert opinion, but is outside the scope of this study.

{\bf{\emph{Rea\_B}}} is the Statlog (German Credit Data) dataset available from the UCI Machine Learning Repository. \emph{Rea\_B} contains $1000$ credit card applications.
It is composed of $20$ attributes describing the status of the applicants pertaining to their credit risk.
Before issuing a credit card, banks would determine if it could be fraudulent based on the features in the data.
Nevertheless, no screening process is perfect, and given a large number
of applications, applications will require retrospective audits to determine whether specific applications should be canceled.
Thus, alerts in this setting
aim to indicate potential fraud and a subset of such alerts are chosen for a time consuming auditing process.
Leveraging the provided features, we define $5$ alert types, which are triggered by the specific combinations of attribute values and the purpose of application.
The $8$ selected purposes of application are the ``victims'' in our audit model.
Table \ref{wrap-tab:11} summarizes how alerts are triggered.
In the description field, italicized words represent the purpose of the application, while the other words represent feature values.

We used the $5$ alert categorizations discussed above to label the $1000$ applications with alert types, excluding any that fail to receive a label.
Among these, we randomly selected $100$ applicants who may choose to ``attack'' one of the $8$ purposes of credit card applications, for a total of $800$ possible events.
The benefit vector for the adversary is $[15,15,14,20,18]$ for each of the alert types generated, respectively.
We set the penalty for detection to $20$ and costs for attack and audit were both set to $1$.
Again, to facilitate comparison we set $p_e = 1$ in all cases.

\begin{table}
	\fontsize{7.5}{7.5}\selectfont
	\centering
	\caption{Description of the defined alert types.}\label{wrap-tab:11}
	\begin{tabular}{clccc}
		\toprule
		\hline
		\\[-0.6em]
		\textbf{ID}  & \textbf{Alert type Description} & \textbf{Mean} & \textbf{Std} \\
		\hline
		\\[-0.6em]
		1 & No checking account, \emph{Any purpose} & 370.04 & 15.81 \\[0.1em]
		2 & Checking $< 0$, \emph{New car, Education} & 82.42 & 7.87  \\[0.1em]
		3 & Checking $> 0$, Unskilled, \emph{Education} & 5.13 & 2.08  \\[0.1em]
		4 & Checking $> 0$, Unskilled, \emph{Appliance} & 28.21 & 5.25   \\[0.1em]
		5 & Checking $> 0$, Critical account,  \emph{Business} & 8.31 & 2.96  \\[0.1em]
		\hline
	\end{tabular}
\end{table}

\subsection{Comparison with Baseline Alternatives}

The performance of the proposed audit model was investigated by comparing with several natural alternative audit strategies as baselines.
The first alternative is to randomize the audit order over alert types, which we call \emph{Audit with random orders of alert types}.
Though random, this strategy mimics the reality of random reporting (e.g., where a random patient calls a privacy official to look into alleged suspicious behavior with respect to the use of their EMR).
In this case, we adopt the thresholds out of the proposed model with $\epsilon = 0.1$ to investigate the performance.
The second alternative is to randomize the audit thresholds. We refer to this policy as \emph{Audit with random thresholds}.
For this policy, we assume that 1) the auditor’s choice satisfies $\sum_i b_i \ge B$ and 2) the auditor has the ability to find the optimal audit order after deciding upon the thresholds.
The third alternative is a naive greedy audit strategy, where the auditor prioritizes alert types according to their utility loss (i.e., greater consequence of violations).
In this case, the auditor investigates as many alerts of a certain type as possible before moving on to the next type in the order. 
For our experiments,
when the alert type order is based on the loss of the auditor, which is the benefit the adversary receives when they execute a successful attack. Thus, we refer to this strategy as \emph{Audit based on benefit}.

\begin{figure}[h]
	\centering
	\includegraphics[width=8.5cm]{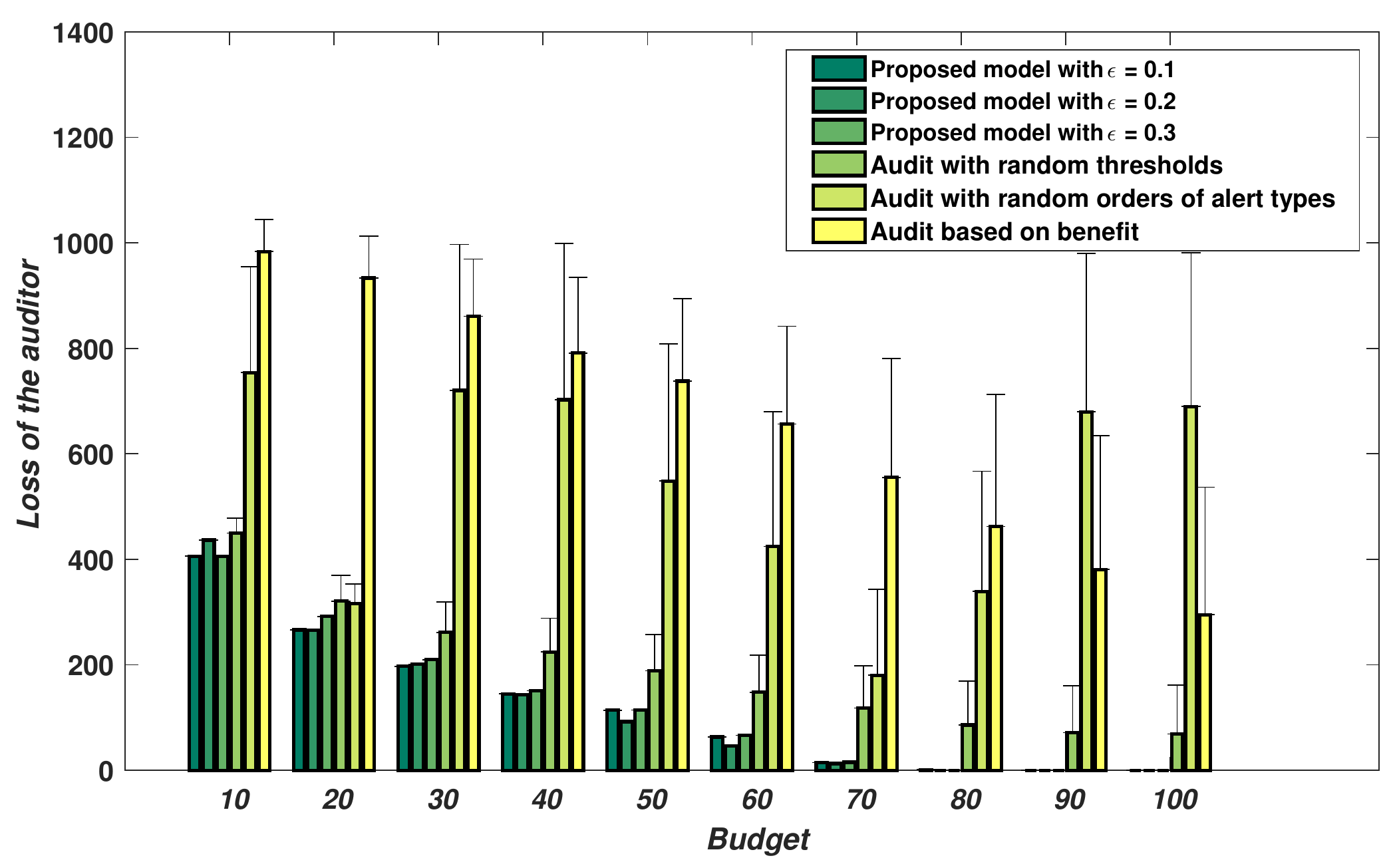}
	\caption{Auditor's loss in the proposed and baseline models in the \emph{Rea\_A} dataset.}
	\label{fig1}
\end{figure}

\begin{figure}[h]
	\centering
	\includegraphics[width=8.5cm]{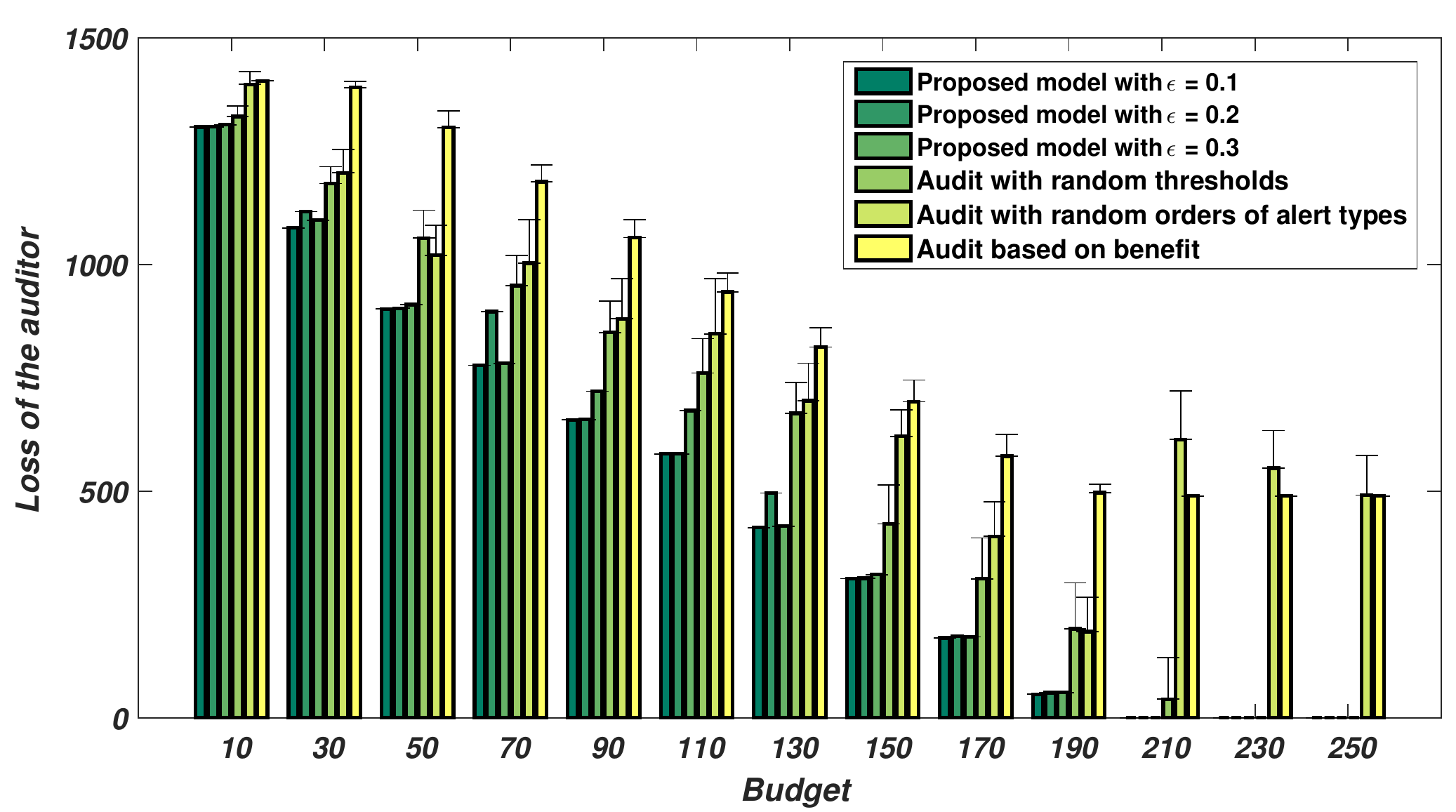}
	\caption{Loss of the auditor in the proposed and alternatives audit model in the \emph{Rea\_B} dataset.}
	\label{fig2}
\end{figure}

The following performance comparisons are assessed over a broad range of auditing budgets. For our model, we present the values of the objective function with three different instances of the step size $\epsilon$ in ISHM: $[0.1, 0.2, 0.3]$.
Figures \ref{fig1} and \ref{fig2} summarize the performance of the proposed audit model and three alternative audit strategies for \emph{Rea\_A} and \emph{Rea\_B}, respectively.

For dataset \emph{Rea\_A}, the range of $B$ was set to $10$ through $100$. The budget of $100$ covers about $1/4$ of the sum of the means of the seven
alert types. In reality, such coverage is quite high.
By applying the proposed audit model, we approximately solve the OAP given
$B$ and $\epsilon$.
For \emph{Audit with random orders of alert types}, we assign the audit thresholds using ISHM with $\epsilon = 0.1$.
The randomization is repeated $2000$ times without replacement. As for \emph{Audit with random thresholds}, we randomly generate the audit thresholds to solve the corresponding LP, which are repeated $5000$ times. For
\emph{Audit based on benefit}, we randomly sample $2000$ instances of $\mathbf{Z}$ based on the distributions of alert types learned from the dataset.

Based on Figure \ref{fig1}, there are several findings we wish to highlight.
First, in our model, as the audit budget increases, the auditor's loss decreases.
At the high end, when $B \ge 90$, the auditor's loss is zero, which, in the VUMC audit setting, implies that all the potential adversaries are deterred from an attack.
This valuation of $B$ is smaller than $1/4$ of the sum of distribution means of all alert types. The reason for this phenomenon stems from the fact that when the audit budget increases, the audit model finding better approximations of the optimal audit thresholds, which, in turn, enables the auditor to significantly limit the potential gains of the adversaries.
Second, our proposed model significantly outperforms all of the baselines.
Third, even though \emph{Audit with random orders of alert types} uses approximated audit thresholds, the auditor's loss is substantially greater than our proposed approach. 
However, the auditor's loss for the alternatives approach ours when $B=20$.  This is because the thresholds are $[0,0,0,7,0,11,8]$, such that the audit order is less of a driver than in other situations.
Fourth, \emph{Audit based on benefit} tends to have very poor performance compared to other policies. This is because when the audit order is fixed (or is predictable), adversaries have greater evasion ability and attack more effectively.
Fifth, \emph{Audit with random thresholds} tends to outperform the other baselines, but is still significantly worse than our approach. The is because
the auditor has the ability to search for the optimal audit policy, but the thresholds are randomly assigned such that they are hampered in achieving the best solution.

For the 
credit card application scenario, Figure \ref{fig2} compares the auditor's loss in our heuristics and the three baselines. For dataset \emph{Rea\_B}, the range for $B$ is $10$ to $250$ with a
step size of $20$.
As expected, as the budget increases, the auditor sustains a decreasing average loss.
It can be seen that the proposed audit model significantly outperforms the alternative baselines. Specifically, as the auditing budget increases, the auditor's loss trends towards, and becomes, $0$ in our approach.  This means that the attackers are completely deterred.
For the alternatives, as before, \emph{Audit with random thresholds} outperforms other strategies.
And, just as before, the strategy that greedily audits alert types (in order of loss) tends to perform quite poorly.

%% file: II_relatedwork.tex
\section{Related Work}
\label{sec:related}

The development of computational methods for raising and subsequently managing alerts in database systems is an active area of research.  In this section, we review recent developments that are most related to our investigation.

\paragraph{Alert Frameworks}

Generally speaking, there are two main categories by which alerts are generated in a TDMT: 1) machine learning methods -- which measure the distance from either normal or suspicious patterns \cite{mathew2010data, kamra2009survey, kamra2008detecting}, and 2) rule-based approaches -- which flag the occurrences of predefined events when they are observed \cite{cook2004rule, samu2002database, ben2008system}.  Concrete implementations are often tailored to distinct application domains.

In the healthcare sector, methods have been proposed to find misuse of EMR systems. Boxwala et at. \cite{boxwala2011using} treated it as a two-label classification problem and trained support vector machines and logistic regression models 
to detect suspicious accesses.  Given that not all suspicious accesses follow a pattern, various
techniques have been developed to determine the extent to which an EMR user \cite{chen2012detecting} or their specific access \cite{chen2012specializing} deviated from the typical collaborative behavior.
By contrast, Fabbri et al. \cite{fabbri2013explaining,fabbri2013select,fabbri2011explanation} designed an explanation-based auditing mechanism which generates and learns typical access patterns from an expert-, as well as data-driven, view. EMR access events by authenticated employees can be explained away by logical relations (e.g., a patient scheduled an appointment with a physician), while the residual can trigger alerts according to predefined rules (e.g., co-workers) or simply fail to have an explanation. The remaining events are provided to privacy officials for investigation; however, in practice, only a tiny fraction can feasibly be audited due to the resource limitation.

In the financial sector, fraud detection \cite{ngai2011application} in credit card applications assists banks in mitigating their losses and protecting consumers \cite{delamaire2009credit}. Several machine learning-based \cite{bhattacharyya2011data} models have been developed to detect fraud behavior. Some of the notable models include hidden Markov models\cite{srivastava2008credit}, neural networks \cite{chan1999distributed}, support vector machines \cite{chen2006new}, etc. Rule-based techniques were also integrated into some detection frameworks~\cite{brause1999neural, sternberg1997using, syeda2002parallel, yeh2009comparisons}. While these methods trigger alerts for investigators, they result in a significant number of false positives---a problem which can be mitigated through alert prioritization schemes.

\paragraph{Alert Burden Reduction}
Various methods have been developed to reduce alert magnitude generated in database systems. Many focus on reducing redundancy and clustering alerts based on their similarity. In particular, a cooperative module was proposed for intrusion detection, which implemented functions of alert management, clustering and correlation \cite{cuppens2002alert}. Xiao et al. proposed a multilevel alert fusion model to abstract high-level attack scenarios to reduce redundancy \cite{xiao2008alert}.  As an alternative, fuzzy set theory was applied by Maggi et al. to design robust alert aggregation algorithms \cite{maggi2009reducing}. Also, a fuzzy-logic engine to prioritize alerts was introduced by Alsubhi et al. by rescoring alerts based on a few metrics \cite{alsubhi2012fuzmet}. Njogu et al. built a robust alert cluster by evaluating the similarity between alerts to improve the quality of those sent to analysts \cite{njogu2010using}.
However, none of these approaches consider the impact of alert aggregation and prioritization on decisions by potential attackers, especially as the latter may choose attacks that circumvent the prioritization and aggregation mechanisms.

\paragraph{Security Games}

Our general model is related to the literature on Stackelberg security games~\cite{kiekintveld2009computing}, where a single defender first commits to a (possibly randomized) allocation of defense resources, while the attacker chooses an optimal attack in response based on observation.
Such models have been applied in a broad variety of security settings, such as airport security~\cite{pita2008deployed}, coast guard patrol scheduling~\cite{an2013deployed}, and even for preventing poaching and illegal fishing~\cite{fang2017paws}.
However, models used in much of this prior work are specialized to physical security, and do not readily generalize to the problem of prioritizing alerts for auditing.
This is the case even for several efforts specifically dealing with \emph{audit games}~\cite{blocki2013audit,blocki2014audit}, which abstract the problem into a set of targets that could be attacked, 
so that the structure of the model remains essentially identical to physical security settings.
In practical alert prioritization and auditing problems, in contrast, a crucial consideration is that there are many potential attackers, and many potential victims or modes of attack for each of these.
Moreover, auditing policies involve recourse actions where the specific alerts audited depend on the realizations of alerts of various types.
Since alert realizations are stochastic, this engenders complex interactions between the defender and attackers, and results in a highly complex space of prioritization policies for the defender. 
In an early investigation on alert prioritization, it was assumed that 1) the identity of a specific attacker was unknown and 2) an exhaustive auditing strategy across alert types of a given order would be applied \cite{laszka2017game}.  These assumptions were relaxed in the investigation addressed by our current study.

Recently, the problem of assigning alerts to security analysts has been introduced~\cite{ganesan2016dynamic}, with a follow-up effort casting it within a game theoretic framework~\cite{Schlenker2017donot}.
The two key limitations addressed by our framework are: 1) it considers only single attacker, whereas auditing decisions in the context of access control policies commonly involve many potential attackers, with most never considering the possibility of an attack; 2) it assumes that the number of alerts in each category is known a priori to both the auditor and attacker.
In practice, alert counts by category are stochastic and can exhibit high variance.

%% file: VII_discussion.tex
\section{DISCUSSION AND CONCLUSIONS}
\label{sec:discussion}

TDMTs are usually deployed in database systems to address a varierty of attacks that originate from within and beyond an organization. However, an overwhelming alert volume is far beyond the capability of auditors with limited resources. Our research illustrates that
policy compliance auditing, as a significant component of database management, can be improved by prioritizing which alerts to focus on via a game theoretic framework, allowing auditing policies to make best use of limited auditing resources while simultaneously accounting for strategic behavior of potential policy violators.
This is notable because
auditing is critical to a wide range of management requirements, including privacy breach and financial fraud investigations.  As such, this model and the effective heuristics we offer in this study
fill a major gap in the field.


There are several limitations of our approach that we wish to highlight as opportunities for future investigations.
First, there are limitations to the parameterization of the game.
One notable aspect is that we assumed that the game has a zero-sum property.
Yet in reality, this may not be the case. For example, an auditor is likely to be concerned less about the cost incurred by an adversary for executing an attack and more concerned about the losses that arise from successful violations 
%
Additionally,
while our experiments show the proposed audit model outperforms 
natural alternatives, it is unclear how sensitive this result is to parameter variations. Thus, more investigation is needed in the next step.

A second set of limitations stems from the assumptions we rely upon.
In particular, we assumed that each attack is instantaneous, which turned the problem into a one-shot two-stage game.
However, attacks in the wild may require multiple cycles to fully execute, such that the auditor may be able to capture the attacker before they complete their exploit.
%
Furthermore, our model is predicated on an environment in which the auditor has complete knowledge, including the identities, about the set of potential adversaries.

A third limitation is in the economic premise of the attack.
Specifically, we expected the interaction between the auditor and adversaries as fully rational. In reality, adversaries may be bounded in their rationality, and an important extension would be to generalize the model consider such behavior.

%% file: IX_acknowledge.tex
\section{Acknowledgement}
This work was supported, in part by grant R01LM10207 from the National Institutes of Health, grant CNS-1526014, CNS-1640624, IIS-1649972 and IIS-1526860 from the National Science Foundation, grant N00014-15-1-2621 from the Office of Naval Research and grant W911NF-16-1-0069 from the Army Research Office.

%% file: X_appendix.tex
\appendix
\section{Appendix}

\textsc{Proof of Theorem \ref{nptheorem}}

\begin{proof}
	We reduce from the 0-1 Knapsack problem defined by Definition \ref{Knapsack}.
	
	We begin by constructing a special case of the auditing problem and work with the decision version of optimization Equation~\ref{E:auditProblem}, in which we decide whether the objective is below a given threshold $\theta$.
	First, suppose that $Z_t = 1$ for all alert types $t \in T$ with probability 1.
	Since the set of orders is a singleton, the probability distribution over orders $p_{\pmb{o}}$ is not relevant.
	Consequently, it suffices to consider $b_t \in \{0,1\}$ for all $t$, and the actual order over types is not
	relevant because $Z_t = 1$ for all types. Consequently, we can choose $\mathbf{b}$ to select an arbitrary subset of types to inspect subject to the budget constraint $B$ (i.e., type $t$ will be audited iff $b_t = 1$).
	Thus, the choice of $\mathbf{b}$ is equivalent to choosing a subset of alert types $A \subseteq T$ to audit.
	
	Suppose that $\mathbf{V} = T$, and each victim $v \in \mathbf{V}$ deterministically triggers some alert type $v \in \mathbf{V} = T$ for any attacker $e$.
	Let $M(\left \langle e,v \right \rangle) = C(\left \langle e,v \right \rangle) = 0$ for all $e \in \mathbf{E}, v \in \mathbf{V}$, and suppose that for every $e$, there is a unique type $t(e)$ with $R(\left \langle e, v \right \rangle) = 1$ if and only if $v = t(e)$ and 0 otherwise.
	Then $\max_v U_a(\mathbf{o}, \mathbf{b}, \left \langle e, v \right \rangle) = 1$ if and only if $b_{t(e)} = 0$ (i.e., alert type $t(e)$ is not selected by the auditor) and 0 otherwise.
	Finally, we let $p_e = 1$ for all $e$.\footnote{While this is inconsistent with our assumption that attackers constitute only a small portion of the system users, we note that this is only a tool for the hardness proof.}
	
	For the reduction, suppose we are given an instance of the 0-1 Knapsack problem.
	Let $T = I$, and for each $i \in I$, generate $v_i$ attackers with $t(e) = i$.
	Thus, $v_i = |\{e : t(e) = i\}|$.
	Let $C_i = w_i$ be the cost of auditing alerts of type $i$, and let $B = W$.
	Define $\theta = |\mathbf{E}| - K$.
	Now observe that the objective in Equation~\ref{E:auditProblem} is below $\theta$ if and only if $\min_{\mathbf{b}}\sum_{t : b_t = 0} v_t \le \theta$, or, equivalently, if there is $R$ such that $\sum_{t \in R} v_t \ge K$.
	Thus, the objective of Equation~\ref{E:auditProblem} is below $\theta$ if and only if the Knapsack instance has a subset of items $R \subseteq I$ which yield $\sum_{i \in R} v_i \ge K$, where $R$ must satisfy the same budget constraint in both cases.
\end{proof}